\def\year{2019}\relax
\documentclass[letterpaper]{article} 
\usepackage{aaai19}  
\usepackage{graphicx}
\usepackage{booktabs}
\newcommand{\citet}[1]{\citeauthor{#1}~\shortcite{#1}}
\newcommand{\citep}{\cite}

\usepackage{times}  
\usepackage{helvet}  
\usepackage{courier}  
\usepackage{url}  
\usepackage{graphicx}  
\usepackage{amsfonts, amsmath, amsthm, bm, amssymb}

\newcommand{\xv}{\mathbf{x}}
\newcommand{\yv}{\mathbf{y}}

\ifx\BlackBox\undefined
\newcommand{\BlackBox}{\rule{1.5ex}{1.5ex}}  
\fi
\ifx\QED\undefined
\def\QED{~\rule[-1pt]{5pt}{5pt}\par\medskip}
\fi
\ifx\proof\undefined
\newenvironment{proof}{\par\noindent{\em Proof:\ }}{\hfill\BlackBox\\[.0mm]}
\fi
\ifx\theorem\undefined
\newtheorem{theorem}{Theorem}
\fi
\ifx\example\undefined
\newtheorem{example}{Example}
\fi
\ifx\property\undefined
\newtheorem{property}{Property}
\fi
\ifx\lemma\undefined
\newtheorem{lemma}{Lemma}
\fi
\ifx\proposition\undefined
\newtheorem{proposition}{Proposition}
\fi
\ifx\remark\undefined
\newtheorem{remark}{Remark}
\fi
\ifx\corollary\undefined
\newtheorem{corollary}{Corollary}
\fi
\ifx\definition\undefined
\newtheorem{definition}{Definition}
\fi
\ifx\conjecture\undefined
\newtheorem{conjecture}{Conjecture}
\fi
\ifx\axiom\undefined
\newtheorem{axiom}[theorem]{Axiom}
\fi
\ifx\claim\undefined
\newtheorem{claim}[theorem]{Claim}
\fi
\ifx\assumption\undefined
\newtheorem{assumption}{Assumption}
\fi

\begin{document}
%
\title{The Level Weighted Structural Similarity Loss: A Step Away from MSE }
\author{Yingjing Lu\\
Carnegie Mellon University\\
yingjinl@andrew.cmu.edu\\
}
\maketitle
\begin{abstract}
The Mean Square Error (MSE) has shown its strength when applied in deep generative models such as Auto-Encoders to model reconstruction loss. However, in image domain especially, the limitation of MSE is obvious: it assumes pixel independence and ignores spatial relationships of samples. This contradicts most architectures of Auto-Encoders which use convolutional layers to extract spatial dependent features. We base on the structural similarity metric (SSIM) and propose a novel level weighted structural similarity (LWSSIM) loss for convolutional Auto-Encoders. Experiments on common datasets on various Auto-Encoder variants show that our loss is able to outperform the MSE loss and the Vanilla SSIM loss. We also provide reasons why our model is able to succeed in cases where the standard SSIM loss fails.   
\end{abstract}

\noindent The MSE has several advantages: it is convex, differentiable, and fast to compute. In the image domain, the MSE has been a great reconstruction loss metric for Auto-Encoders (AE) \citep{hinton2006reducing}. Even though the MSE has proved to be the most successful loss for AEs, its major feature, measuring error of each dimension independently, does not align with the assumption in the actual model as we point out in the case of convolutional Auto-Encoders (C-AE). A C-AE uses its sliding convolutional filters to extract features such as ears or eyes, which requires to sample an area rather than individual pixels to encode meaningful information. Thus C-AE assumes that pixels are depending on its surrounding pixels to express information which is not aligning with the assumption of MSE. Another problem with using the MSE as reconstruction loss is that it measures absolute pixel differences rather than the structural difference which is more aligned with human's perception. 

Existing works that seek alternative reconstruction loss mainly utilize the SSIM proposed by \citep{wang2004image}. However, despite the fact that it has shown its strength in measuring the image similarity, it has some limitations including not as effective in RGB images as it does in grey-scale images, and not being sensitive to luminance. Some works seek to alleviate its limitations by combining the SSIM score with the Mean Absolute Error (MAE) or the MSE \citep{zhao2017loss}, but those formulations lack straightforward explanations. 

In the following sections we show SSIM's limitations by providing mathematical intuitions and empirical results. We then propose LWSSIM, a revised SSIM loss, and provide intuitive justifications for our formulation and experiment results.

\section{The SSIM and Its Limitations}

The SSIM is originally constructed as an image quality measure with respect to the human perception rather than absolute differences measured by metrics such as the MSE or Mean Absolute Error (MAE).

To express the SSIM formally, according to \citep{wang2004image}, consider a pair of images $\{x, y\}$ of sizes $m \times n$, we want to measure three aspects of similarities according to human perception: luminance($l(x,y)$), contrast($c(x,y)$), and structure($s(x,y)$). Those are quantified according to the summary of relative measures including mean, variance, and co-variance measured under sliding windows of size $\xi \times \xi$ with step size of 1 on both horizontal and vertical directions. 

For each sliding window, we measure base quantity for images $x$ and $y$ respectively. Then each perception sub-function is computed as follows:
\begin{equation}
    l(x,y) = \frac{2\mu_x\mu_y + C_1}{\mu_x^2 + \mu_y^2 + C_1}
\end{equation}
\begin{equation}
    c(x,y) = \frac{2\sigma_x\sigma_y + C_2}{\sigma_x^2 + \sigma_y^2 + C_2}
\end{equation}
\begin{equation}
    s(x,y) = \frac{\sigma_{xy} + C_3}{\sigma_x\sigma_y + C_3}
\end{equation}

Here $\{C_1, C_2, C_3\}$ are constants less than 1 to balance potential zero division issue. Usually, $C_3 = \frac{1}{2}C_2$. 

To enforce independence among those measures, the final SSIM is constructed as the product of those metrics with exponential constant weights $\{\alpha, \beta, \gamma \}$ as:
\begin{equation}
    SSIM( x, y ; \xi ) = l(x,y)^{\alpha} \cdot c(x,y)^{\beta} \cdot s(x,y)^{\gamma}
\end{equation}

Notice that the luminance function is different from contrast and structure measures in that it uses mean to measure relative difference in luminance as oppose to absolute luminance difference. Mathematically, mean measures can output same value despite the range of pixel value in our measure window. In other word, the SSIM is not able to discriminate between a picture with high light-dark variance and a picture with more consistent light and dark if the two pictures have same mean luminance value. Further, since the SSIM formulation uses multiplication between different sub-functions thus encourages independence among those factors, luminance's limitation is hard to be compensated by contrast or structure measures. 

This is not a problem when measuring similarities between two images because the values of the two images are fixed. However, this is problematic when we want to use the SSIM as a loss function to guide the image reconstruction. Models with the SSIM as loss would eventually not able to learn to discard images generated with low luminance variance if other metrics are optimal. 

\section{Proposed Structural Similarity Loss}
We aim to adjust the existing SSIM metric to increase information acquisition in two folds. First, we replace the multiplication between luminance and the other with addition, and thus is able to remove independence between the three sub-functions to allow variance and co-variance to compensate mean. Secondly, we calculate the score under different filter size and calculate the weighted average among different level of filter sizes. This aims to alleviate the problem that the luminance metric is over "averaged" due to its large filter area. Different sized filters also allow the loss to capture different levels of localized information.

Formally, consider a pair of images $\{x, y\}$ of size $m \times n$ where $x$ is sampled from true distribution $p(x)$ and $y$ is sampled from the Auto-Encoder network using a corresponding $x$: $y = h_{\theta \phi}(x)$ where ${\theta, \phi}$ denotes parameters of encoder and decoder network respectively. For the $i^{th}$ level of similarity measure, we apply sliding window of different size $\xi_i \times \xi_i$ that moves pixel-by-pixel over the entire image. 
Then for each $\xi_i$ we have:
\begin{align}
    LWSSIM( x, y ; \xi ) = l(x,y)^{\alpha} + c(x,y)^{\beta} \cdot s(x,y)^{\gamma}
\end{align}
And for $I$ levels of different filter sizes we calculate the weighted sum with hyperparameters $\lambda_i$
\begin{align}
    LWSSIM( \xv, \yv ) = \frac{1}{I}\sum_{i}\lambda_i LW\-SSIM( \xv, \yv; \xi_i)
\end{align}

\section{Experiments \& Results}
We applied our loss to replace the MSE part of loss functions in trending AE models including vanilla Auto-Encoder, Variational Auto-Encoder (VAE) \citep{kingma2013auto}, and MMD AE \citep{tolstikhin2017wasserstein}. All of the aforementioned models were implemented with convolutional layers in encoders and decoders. Observing figure 1, compared to images generated from models trained with MSE loss, we can see that the original SSIM loss enables the trained AE to generate less blurry images but with less vibrant colors resulted from discussed luminance issue across three channels. Our LWSSIM loss allows AE to generate images with higher levels of detail while preserving the color vibrancy. From quantative perspective, we see that the LWSSIM loss allows the model to generate images not only with higher structural similarity scores but also with lower MSE values most of the times, indicating that the LWSSIM is a good candidate to ensure both structural and absolute similarity. 

\begin{table}[t]
\caption{Reconstruction Quality Results on Celeb-A 64}
\label{sample-table}
\begin{center}
\begin{small}
\begin{sc}
\begin{tabular}{lccccr}
\toprule
Model & SSIM & MSE \\
\midrule
AE (MSE) & 0.80 & 0.0042 \\
AE (SSIM) & 0.86 & 0.0057 \\
\textbf{AE (LWSSIM)} & \textbf{0.89} & \textbf{0.0054}\\
\midrule
VAE (MSE) & 0.53 & 0.0235\\
VAE (SSIM) & 0.60 & 0.0244\\
\textbf{VAE (LWSSIM)} & 0.59 & \textbf{0.0206}\\
\midrule
MMD AE (MSE) & 0.81 & 0.0034\\
MMD AE (SSIM) & 0.86 & 0.0062\\
\textbf{MMD AE (LWSSIM)} & \textbf{0.88} & \textbf{0.0033}\\
\bottomrule
\end{tabular}
\end{sc}
\end{small}
\end{center}
\vskip -0.1in
\end{table}

\begin{figure}
  \centering
  \includegraphics[width=\linewidth]{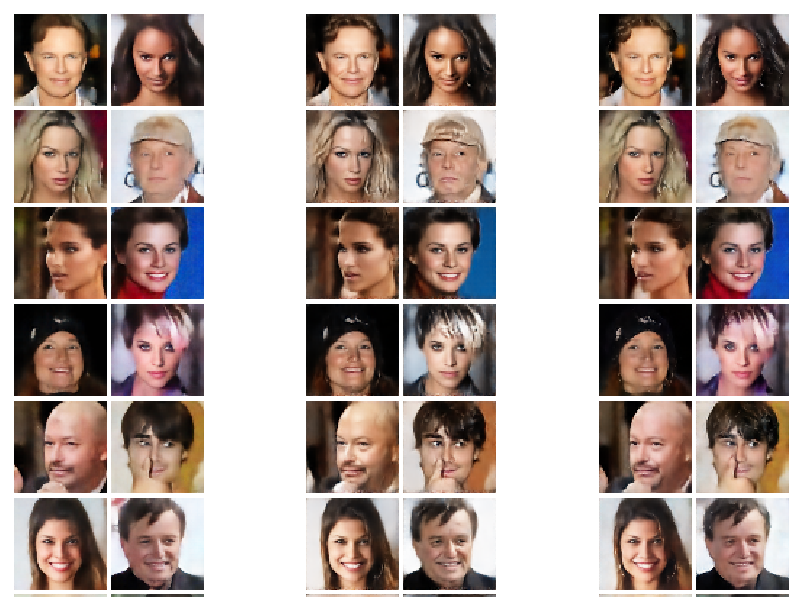}
  \caption{MMD AE reconstruction, MSE(left), SSIM(mid), LWSSIM(right). Notice the color loss in SSIM pics.}
\end{figure}

\bibliographystyle{aaai}
\bibliography{macro}


\end{document}